\documentclass{bmvc2k}
\usepackage{times}
\usepackage{epsfig}
\usepackage{graphicx}
\usepackage{amsmath}
\usepackage{amssymb}
\usepackage{booktabs, adjustbox, nccmath}
\usepackage[symbol]{footmisc}

\title{Author Guidelines for the\\ British Machine Vision Conference}

\addauthor{Julia Gong}{jxgong@cs.stanford.edu}{1}
\addauthor{F. Christopher Holsinger}{holsinger@stanford.edu}{1}
\addauthor{Serena Yeung}{syyeung@stanford.edu}{1}

\addinstitution{
 Stanford University\\
 Stanford, California, USA
}

\runninghead{Gong \etal}{Weakly-Supervised Visual Warping for Single-Shot VOS}

\def\etal{\emph{et al}\bmvaOneDot}

\makeatletter
\newcommand\footnoteref[1]{\protected@xdef\@thefnmark{\ref{#1}}\@footnotemark}
\makeatother

\begin{document} 

\title{FlowVOS: Weakly-Supervised Visual Warping for Detail-Preserving and Temporally Consistent Single-Shot Video Object Segmentation}

\author{Julia Gong, F. Christopher Holsinger, Serena Yeung\\
Stanford University\\
Stanford, California, USA\\
{\tt\small jxgong@cs.stanford.edu, holsinger@stanford.edu, syyeung@stanford.edu}
}

\maketitle

\vspace{-1.5em}
\begin{abstract}
    We consider the task of semi-supervised video object segmentation (VOS). Our approach mitigates shortcomings in previous VOS work by addressing detail preservation and temporal consistency using visual warping. In contrast to prior work that uses full optical flow, we introduce a new foreground-targeted visual warping approach that learns flow fields from VOS data. We train a flow module to capture detailed motion between frames using two weakly-supervised losses. Our object-focused approach of warping previous foreground object masks to their positions in the target frame enables detailed mask refinement with fast runtimes without using extra flow supervision. It can also be integrated directly into state-of-the-art segmentation networks. On the DAVIS17 and YouTubeVOS benchmarks, we
    outperform state-of-the-art offline methods that do not use extra data, as well as many online methods that use extra data. Qualitatively, we also show our approach produces segmentations with high detail and temporal consistency. 
\end{abstract}

\section{Introduction}
\vspace{-0.2em}

Video object segmentation (VOS) has become an increasingly studied task in the computer vision community. The goal of VOS is to label each pixel of each frame of a video with a corresponding class---either one of potentially several foreground objects, or the background. In particular, the semi-supervised inference setting of this task provides the ground-truth segmentation mask for the first video frame, and methods aim to segment these objects for all subsequent frames. This task is difficult because objects in motion can move and deform in different ways, not to mention additional challenges such as camera motion and occlusions.

\begin{figure}[t]
\centering
\includegraphics[width=0.6\linewidth]{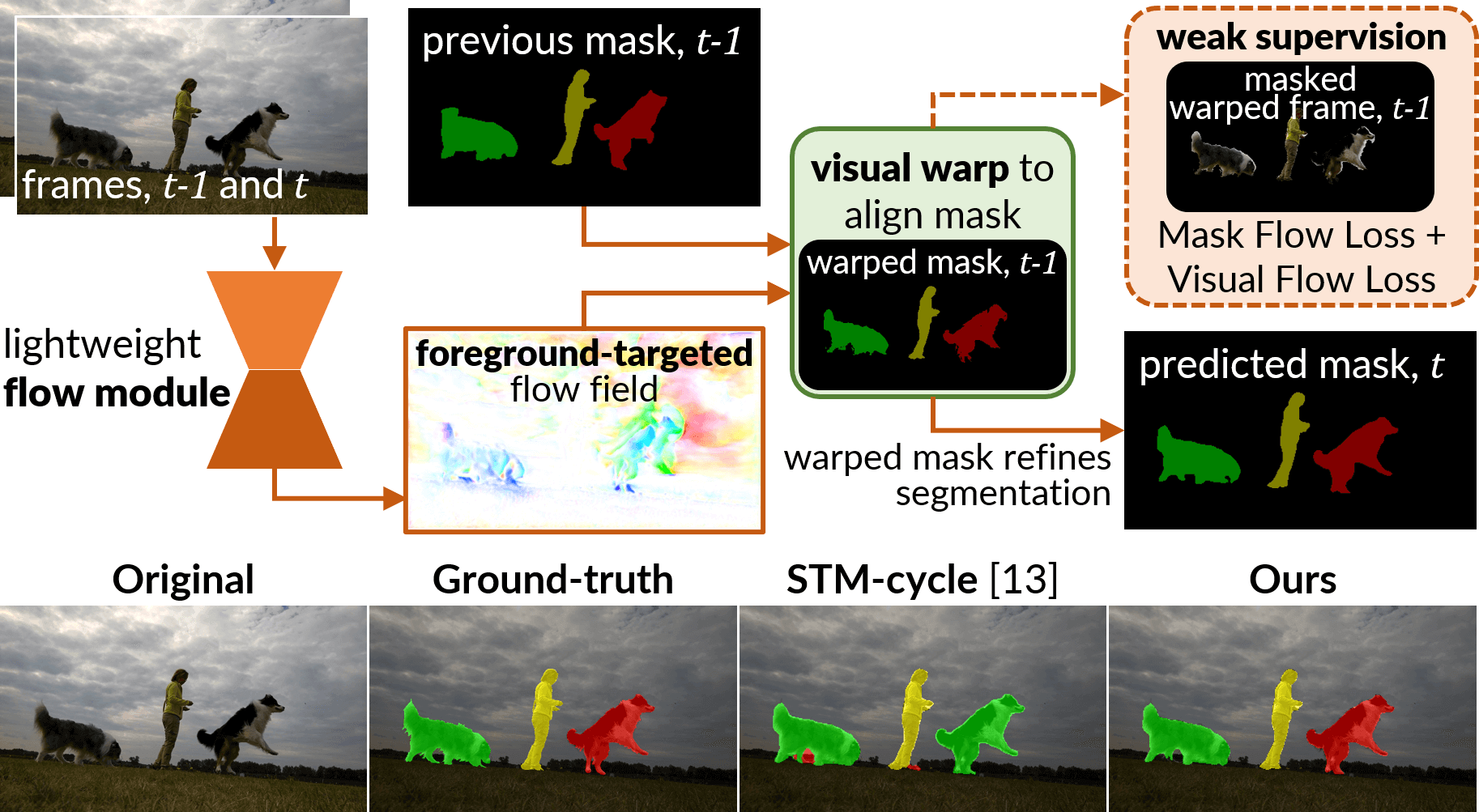}
\vspace{-0.5em}
   \caption{We introduce a weakly-supervised visual warping approach for VOS that improves detail and temporal consistency. Our lightweight flow module learns to regress a foreground-targeted flow field that warps the previous mask from time $t-1$ to align it to frame $t$. To learn detailed flow fields, we directly exploit the VOS data using two weakly-supervised losses, rather than learning the cumbersome general optical flow task. We show that our warped masks effectively refine the final segmentations. Our method can be easily integrated into state-of-the-art segmentation models, does not require extra data, and has fast frame rates.}
\label{fig:pullfig}
\vspace{-1em}
\end{figure}

Many deep learning-based methods have been proposed to tackle this problem. Recent state-of-the-art offline methods~\cite{jiang2019stm, li2020stmcycle} have used a memory bank of encoded previous frames and masks, which is queried when segmenting later frames.
The advantage of these approaches is that the learned latent spaces robustly encode higher-level features of the target objects from previous frames; however, they lack the detail to propagate fine features and movements across consecutive frames and thus can suffer from temporal inconsistency.

Our work's key insight is to mitigate these issues of detail preservation and temporal consistency using \textit{visual warping}, which captures small deviations between video frames. As shown in Figure~\ref{fig:pullfig}, our method warps the object masks from previous frames toward the target frame to add detail and temporal consistency to the final segmentation. To learn these deltas between frames, we introduce a weakly-supervised flow module that can be easily used with state-of-the-art segmentation networks. Some early VOS works employed traditional optical flow methods to perform visual warping~\cite{tsai2016video, perazzi2017learning}, but they use highly time-costly online optical flow optimization methods. 
More recent VOS works~\cite{cheng2017segflow, luiten2018premvos, lin2019agss} incorporate warping by using state-of-the-art deep learning optical flow estimation networks like FlowNet 2.0~\cite{ilg2017flownet2}. However, these networks have high computational cost and require extensive pretraining on supervised data focused on the full optical flow task~\cite{dosovitskiy2015flownet, mayer2016things3d}. Their ability to predict detailed motions also degrades significantly even with small speed increases.

In this work, we address these shortcomings by proposing a novel visual warping approach for VOS. 
Unlike prior works that approach visual warping using standalone, pretrained optical flow methods, we do not 
predict traditional optical flow. Instead, we directly train on the VOS data of interest to learn an offline flow module for VOS-specific visual warping. Our foreground-targeted approach focuses on aligning previous foreground objects to their new positions. Specifically, we introduce two weakly-supervised flow losses that enforce pixel-level consistency between warped previous masks and frames and the target masks and frames.
These train the flow module to capture small changes between timesteps in pixel-wise flow fields, which warp previous object masks to propagate detail to the final prediction. 
Moreover, since we do not learn the general optical flow task, our flow module can stay lightweight for fast frame rates, and by training directly on the VOS data, we do not need any supervised flow data. As such, our method can generalize to diverse data not studied by traditional optical flow techniques, and more generally to object-focused scenes.

On the two major VOS benchmarks, DAVIS17~\cite{perazzi2017learning} and YouTubeVOS~\cite{xu2018youtube}, our method achieves state-of-the-art performance among offline works that do not use extra training data (e.g. additional datasets, image segmentations, supervised optical flow, or synthetic data). Additionally, we outperform or stay competitive with those that use online learning and extra data. We also qualitatively show that our method achieves greater segmentation detail preservation and temporal consistency. Our contribution can be summarized as follows:
\vspace{-0.5em}
\begin{itemize}
    \setlength\itemsep{0.15em}
    \item We propose a novel foreground-targeted visual warping approach that improves segmentation detail and temporal consistency for VOS. We show that instead of learning traditional optical flow, our flow module jointly learns detail-preserving flow fields by exploiting the target VOS data directly using two weakly-supervised losses.

    \item Our purely offline-learned flow module for VOS is fast and can be easily integrated into state-of-the-art segmentation networks (here, we integrate it into STM-cycle~\cite{li2020stmcycle}).
    
    \item On DAVIS17~\cite{pont-tuset2017davis} and YouTubeVOS~\cite{xu2018youtube}, we achieve state-of-the-art performance among works that do not use online learning nor extra data.
    We also outperform or stay competitive with those that do, while maintaining faster frame rates.
\end{itemize}

\vspace{-1.9em}
\section{Related Work}
\vspace{-0.75em}

\paragraph{Semi-Supervised Video Object Segmentation.} With the success of deep learning, semi-supervised VOS has seen a large number of works in recent years. These works leverage a variety of strategies, including online versus offline optimization, mask propagation, segmentation by tracking, coarse-to-fine refinement, and usage of attention and memory banks.

Semi-supervised VOS methods can be considered \textit{online} or \textit{offline}, where inference includes learning in the former and does not in the latter. Online techniques are often used for mask refinement. OSVOS~\cite{caelles2017osvos} introduced the first deep learning-based online VOS method, which gradually refines the model from segmenting general objects to those in the initial reference mask; OnAVOS~\cite{voigtlaender17onavos} adds an adaptive learning mechanism. MaskTrack~\cite{perazzi2017learning}'s online learning method learns mask refinement from external static images. PReMVOS~\cite{luiten2018premvos} achieves strong performance via coarse-to-fine refinement of object proposals and optical flow, though it is among the most computationally intensive. More recent work also employs online learning on the initial reference mask to refine the prediction~\cite{huang2020template}.
While online learning methods can produce detailed segmentations, their high computational cost causes slow inference frame rates impractical for real-time settings.

In contrast, offline learning methods do not update during inference, generally yielding faster frame rates. Our method lies in this category.
Recently, D3S~\cite{lukezic2020d3s} proposed segmenting frames independently with explicit foreground-background separation, though its temporal consistency drops in multi-object settings. 
To enforce consistency, some offline methods leverage previous frames; S2S~\cite{xu2018youtube} and RVOS~\cite{ventura2019rvos} use recurrent networks, while RGMP~\cite{oh2018rgmp} use Siamese encoders for the previous and current frame. The weakness of these methods is that they do not target specific features across frames. Many methods~\cite{xu2019stcnn, voigtlaender2019feelvos, lin2019agss, jiang2019stm, li2020stmcycle} therefore use attention mechanisms to achieve stronger performance.

One such work, FEELVOS~\cite{voigtlaender2019feelvos}, leverages the initial reference mask for explicit feature matching with subsequent frames. AGSS-VOS~\cite{lin2019agss} further attends over the previous frame and mask, while STCNN~\cite{xu2019stcnn} attends over multi-scale feature maps. STM~\cite{jiang2019stm} achieves even higher performance by introducing an external memory bank and attending over multiple previous frames by querying them using the target frame. State-of-the-art STM-cycle~\cite{li2020stmcycle} adds a cyclic loss to reduce error propagation.
Concurrently, some works~\cite{mei2021transvos, duke2021sstvos} instead use attention via transformers to address pixel spatiotemporal relations and model scalability.
While attention works capture higher-level features of foreground objects, their segmentations often lack detail or fail to propagate fine deformations and movements across frames, even when the object does not change much. Our insight is that explicitly learning the differences between pairs of frames can address these issues. Thus, improving on previous work, we leverage visual warping to add temporal consistency and segmentation detail.

\vspace{-1.6em}
\paragraph{Optical Flow Estimation in VOS.}
Previous VOS works that perform visual warping all use optical flow prediction methods to do so.
Early VOS works such as~\cite{tsai2016video} and MaskTrack~\cite{perazzi2017learning} use traditional online optical flow estimation methods. Later work uses deep learning-based optical flow prediction approaches, such as FlowNet~\cite{dosovitskiy2015flownet}, which pioneered the task as a supervised problem using convolutional neural networks. Using~\cite{dosovitskiy2015flownet}, SegFlow~\cite{cheng2017segflow} jointly learns optical flow and segmentation with deep learning, but requires online learning and learns the full optical flow task, thus requiring supervised flow annotations. Works such as~\cite{tsai2016video, perazzi2017learning, cheng2017segflow} are either online, use online optical flow estimation, or both, rendering them computationally costly and unfit for real-time settings.~\cite{perazzi2017learning, cheng2017segflow} also require extra data for training, and they do not exploit the warped masks themselves to guide the segmentation.

More recent VOS work uses state-of-the-art optical flow prediction networks that outperform FlowNet~\cite{dosovitskiy2015flownet}; 
notably, FlowNet 2.0~\cite{ilg2017flownet2} achieves a significant improvement by stacking multiple convolutional networks end-to-end. While~\cite{ilg2017flownet2} is state-of-the-art, it requires extensive pretraining on several supervised optical flow datasets~\cite{dosovitskiy2015flownet, mayer2016things3d} and is computationally costly to stack on top of a segmentation model. Moreover, without extensive ensembling of multiple networks end-to-end, its ability to predict detailed movements drops. Still, many strong VOS works accept these tradeoffs and use~\cite{ilg2017flownet2}. In particular, recent works PReMVOS~\cite{luiten2018premvos} and AGSS-VOS~\cite{lin2019agss} use the pretrained FlowNet 2.0~\cite{ilg2017flownet2} to predict optical flow, which they use to warp previous masks either for further refinement or to guide attention mechanisms. However, these works suffer from FlowNet 2.0~\cite{ilg2017flownet2}'s accuracy-speed tradeoff; they use the FlowNet 2.0 ensemble at the cost of significant frame rate drops.
Moreover, they rely on the extensive extra flow data used to pretrain FlowNet 2.0, which generates additional complexity and data requirements, while importantly not being foreground object-targeted.

Our approach mitigates these issues in prior work. Specifically, our key insight is to learn a weakly-supervised, foreground-targeted visual warping model for VOS instead of learning general optical flow. 
Even with no extra data and at faster speeds, our approach produces detailed and temporally consistent segmentations and can train directly on the target data.

\vspace{-1.5em}
\section{Methods}
\vspace{-1em}

\begin{figure*}
\centering
\includegraphics[width=\linewidth]{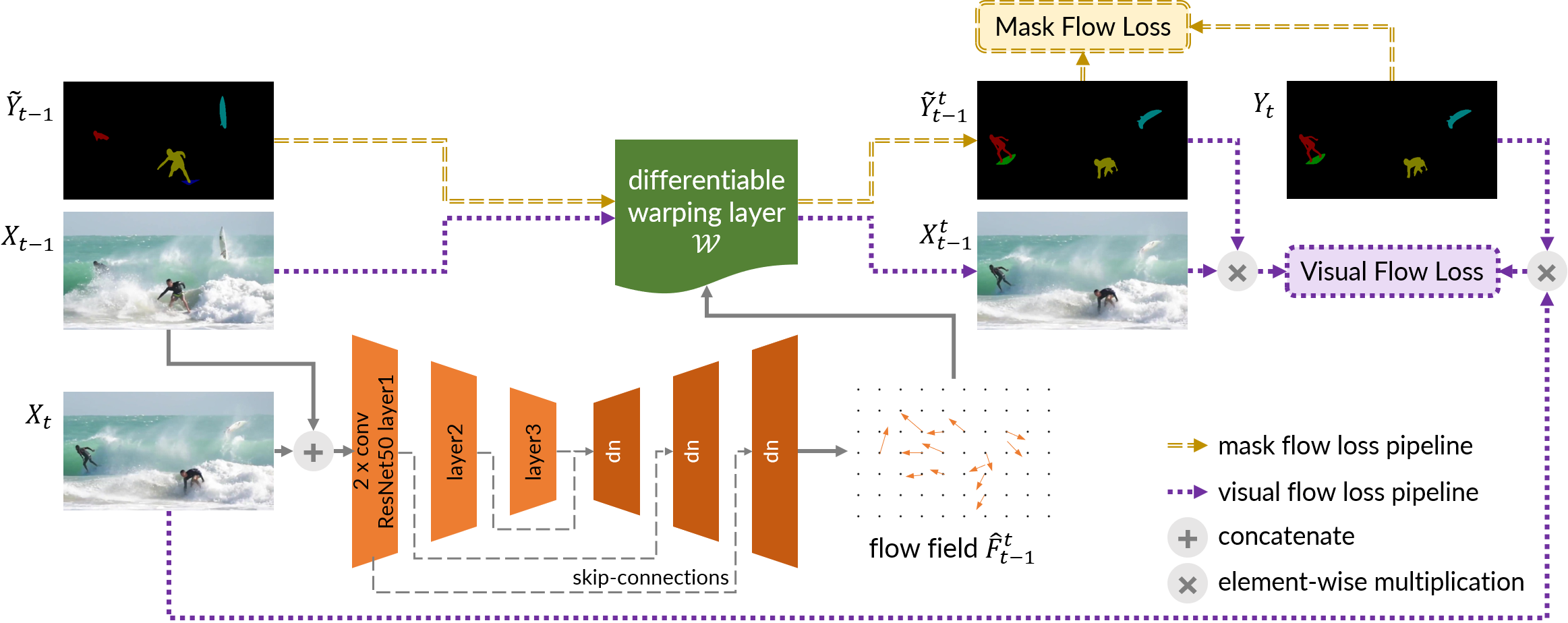}
   \caption{Overview of the proposed flow module. Given a previous and target frame (here shown as $X_{t-1}, X_t$), it regresses a flow field to warp the previous frame and mask to the current ones. The previous mask $\Tilde{Y}_{t-1}$ can be either the ground-truth or predicted mask from timestep $t-1$ in training (Sec.~\ref{sec:training}), but only the predicted mask in inference.
   Two weakly-supervised losses train the flow module: the Mask Flow Loss minimizes the difference between the warped previous mask $\Tilde{Y}_{t-1}^t$ and target mask $Y_t$ (yellow double-dashes), and the Visual Flow Loss minimizes the difference between the warped previous frame $X_{t-1}^t$ masked by $\Tilde{Y}_{t-1}^t$ and the target frame $X_t$ masked by $Y_t$ to eliminate background noise (purple dotted lines).}
\label{fig:flowmodule}
\vspace{-0.5em}
\end{figure*}

Our method tackles semi-supervised video object segmentation (VOS), improving the detail and temporal consistency of prior work. The key contribution of our approach is the weakly-supervised flow module, which learns to regress foreground object-targeted flow fields that warp previous masks toward objects in the target frame to preserve fine detail and temporal consistency. As it need not learn general optical flow, it can remain lightweight. It can be used with any segmentation network to refine masks; we use STM-cycle~\cite{li2020stmcycle} in this work. To train the flow module, we introduce two weakly-supervised losses for flow field regression, which encourage pixel-level consistency between warped previous object masks and frames and the target masks and frames. Unlike prior work~\cite{luiten2018premvos, lin2019agss}, these losses exploit the VOS data to regress object-targeted flows without any extra supervision from ground-truth flow fields. Below, we review the semi-supervised VOS problem definition, describe our model architecture, formalize the two weakly-supervised flow losses, and discuss model training.

We first define notation for semi-supervised VOS. Given a video with $T$ frames, let the $t$-th frame in temporal order be $X_t$ and its corresponding ground-truth mask be $Y_t$, for $t \in [1, T]$.
In training, all ground-truth masks are provided; in inference, only the first ground-truth mask $Y_1$ is provided. The goal of the model is to predict masks $\hat{Y}_t$ for all subsequent frames.

\subsection{Flow Module} \label{sec:flowmodule}
As shown in Figure~\ref{fig:flowmodule}, we introduce a lightweight flow module $\mathcal{F}$ to address detail preservation and temporal consistency. $\mathcal{F}$ is an hourglass network to enable learning features of different scales (Implementation Details in Sec.~\ref{sec:implementation}). Given a pair of frames from the same video, $\mathcal{F}$ generates a flow field that describes object movement between the previous and current frame. 
For a target frame $X_t$, the flow module takes two frames in temporal sequential order, here $\{X_{t-1}, X_t\}$, and an object mask $\Tilde{Y}_{t-1}$ for the previous frame (in training either the ground-truth $Y_{t-1}$ or predicted mask $\hat{Y}_{t-1}$ (Sec.~\ref{sec:training}); in inference only the latter). It outputs a flow field $\hat{F}_{t-1}^t$ that warps $X_{t-1}$ to $X_t$ and $\Tilde{Y}_{t-1}$ to $Y_t$. The flow field has the same spatial dimensions as the video frames, with two channels corresponding to pixel-wise $x$ and $y$ displacements normalized by frame width and height. The flow module's function is thus
\begin{equation} \label{eq:flowregression}
    \hat{F}_{t-1}^t = \mathcal{F}(X_{t-1}, X_t, \Tilde{Y}_{t-1}).
\end{equation}

This flow field $\hat{F}_{t-1}^t$ warps the previous object mask~$\Tilde{Y}_{t-1}$ toward the target mask using the differentiable warping layer $\mathcal{W}$ introduced in Spatial Transformer Networks~\cite{jaderberg2015spatial}, which transforms images using a sampling kernel. The resulting warped previous mask~$\Tilde{Y}_{t-1}^t$ is thus
\begin{equation} \label{eq:warp}
    \Tilde{Y}_{t-1}^t = \mathcal{W}(\Tilde{Y}_{t-1}, \hat{F}_{t-1}^t).
\end{equation}

We train the flow module to regress effective visual warping flow fields with only VOS data and no extra flow data. To do so, we introduce two weakly-supervised losses.

\subsection{Weakly-Supervised Flow Losses}
Our work contributes a novel weakly-supervised approach to regress visual warping flow fields for VOS. 
In contrast to prior works that use heavy pretrained optical flow models, our key insight is to leverage existing VOS data to directly learn flow fields in a weakly-supervised manner using two losses: the Mask Flow Loss (MFL) and Visual Flow Loss (VFL). Both train the flow module to warp previous objects toward the targets by penalizing pixel-level differences after warping. 
The MFL minimizes differences between warped and ground-truth target masks, while the VFL does the same for warped and target video frames.

\textbf{Mask Flow Loss (MFL).} Our goal is to align previous object masks to the target frame. As such, we require the warped previous mask $\Tilde{Y}_{t-1}^t$ (Eq.~\ref{eq:warp}) to be close to the target mask $Y_t$. The MFL minimizes the difference between the warped previous mask and target mask.
It combines the commonly-used cross-entropy and mask IOU losses between these two masks,
\begin{align} \label{eq:maskflowloss}
    \mathcal{L}_{MF} =& \medmath{ \dfrac{1}{|P|} \sum_{u \in P} \left( (1-Y_{t,u}) \log(1-\Tilde{Y}_{t-1,u}^t) + Y_{t,u} \log(\Tilde{Y}_{t-1,u}^t) \right) } \nonumber \\
    &-\lambda \frac{\sum_{u \in P} \min(\Tilde{Y}_{t-1,u}^t, Y_{t,u})}{\sum_{u \in P} \max(\Tilde{Y}_{t-1,u}^t, Y_{t,u})},
\end{align}
where $P$ is the set of pixel coordinates, $Y_{t,u}$ and $\Tilde{Y}_{t-1,u}^t$ are the mask pixel values at coordinate $u$ for the ground-truth and warped previous masks respectively, and $\lambda$ weights the two losses.
This combines their strengths: cross-entropy favors pixel-level accuracy and optimizes more stably, while the IOU loss enforces overall object shape and better handles class imbalances.

\textbf{Visual Flow Loss (VFL).} To warp previous objects to the target frame, we can similarly exploit visual appearance; we thus also require the warped previous frame denoted $X_{t-1}^t$ (achieved via the analogous operation on $X_{t-1}$ using Eq.~\ref{eq:warp}) to be close to the target frame $X_t$. A naive formulation of the VFL may be the pixel-wise mean squared error (MSE) between the warped previous frame and the target frame. However, the disadvantage of this formulation is that video frames can exhibit large amounts of visual noise in the background that is irrelevant to the objects of interest. This could be caused by background activity, camera motion, occlusions, or motion blur, among other factors.

Thus, since we are only concerned with the motion of the foreground objects for the VOS task, a stronger Visual Flow Loss will only take these pixels into consideration to more precisely capture object movement. Specifically, we use the MSE loss between two \textit{masked} frames: the target frame $X_t$ masked with the target mask $Y_t$, and the warped previous frame $X_{t-1}^t$ masked with the warped previous mask $\Tilde{Y}_{t-1}^t$ to obtain (continuing the notation in Eq.~\ref{eq:maskflowloss}):
\begin{equation}
    \mathcal{L}_{VF} = \frac{1}{|P|} \sum_{u \in P} \left((X_{t-1,u}^t) (\Tilde{Y}_{t-1,u}^t) - (X_{t,u}) (Y_{t,u}) \right)^2.
\end{equation}

\subsection{End-to-End Segmentation Method}

\begin{figure}[t]
\centering
\includegraphics[width=0.54\linewidth]{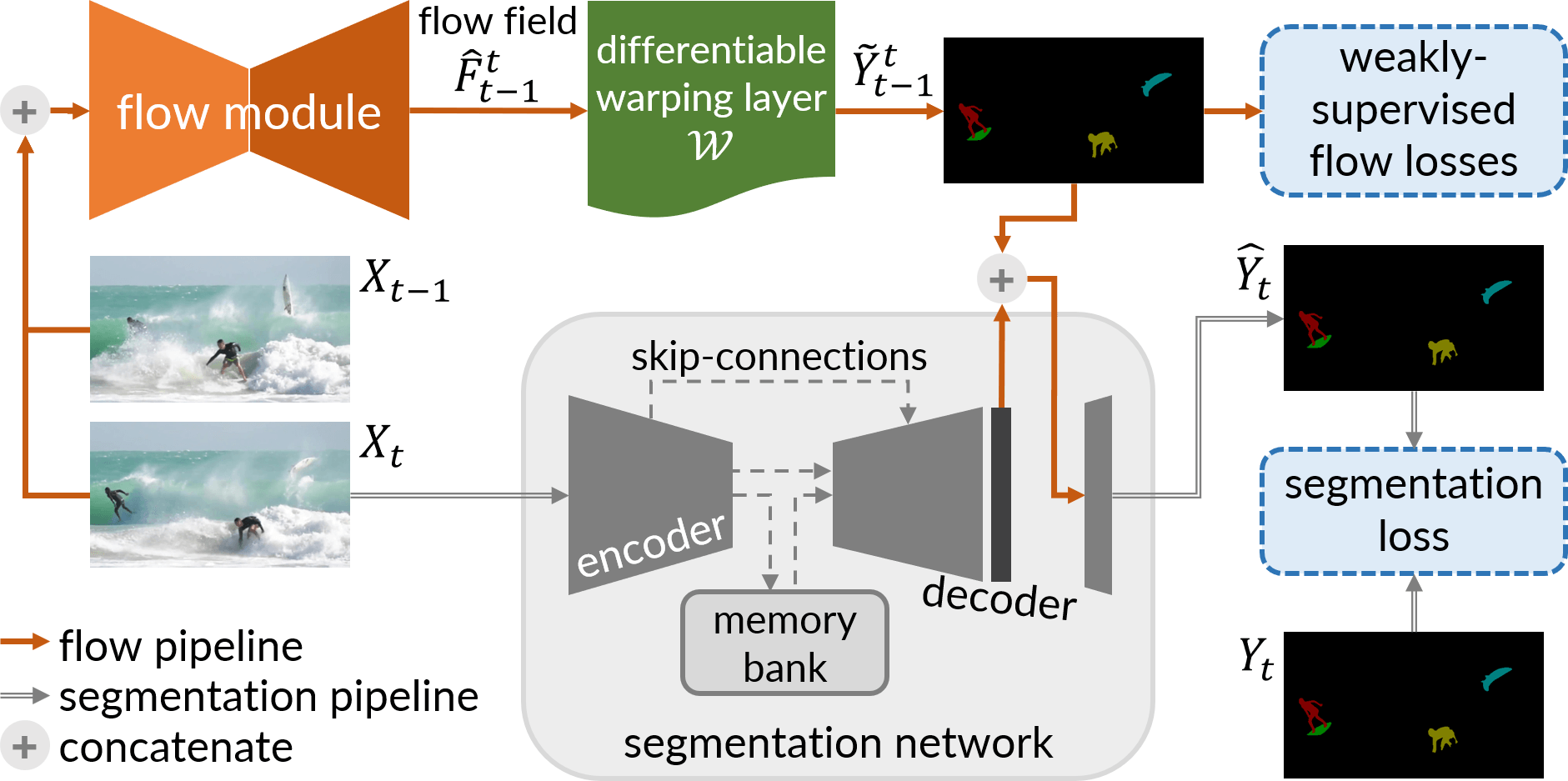}
\vspace{-0.5em}
   \caption{End-to-end FlowVOS method. Our flow module takes the previous and target frames $X_{t-1}, X_t$ as input. We train it to generate a flow field $\hat{F}^t_{t-1}$ that warps a previous mask to align it with the target frame $X_t$ using two weakly-supervised losses. The warped mask $\Tilde{Y}_{t-1}^t$ is concatenated to the segmentation decoder's second-to-last convolutional block feature map to predict the final mask $\hat{Y_t}$. In training, the  weakly-supervised losses are used together with a standard segmentation loss (mask IOU and cross-entropy, as in Eq.~\ref{eq:maskflowloss}).
   }
\label{fig:flowvos}
\vspace{-0.5em}
\end{figure}

Our flow module can be easily integrated into state-of-the-art segmentation networks; here, we present a version of our method that uses STM-cycle~\cite{li2020stmcycle}, which is based on STM~\cite{jiang2019stm}. 
Our end-to-end method, shown in Figure~\ref{fig:flowvos}, uses the flow module's output flow field to refine the final segmentation. To integrate our flow module into the segmentation network, the regressed flow field first warps the previous mask $\Tilde{Y}_{t-1}$ to yield $\Tilde{Y}_{t-1}^t$, as discussed in Eq.~\ref{eq:warp}.
We concatenate this warped previous mask with the output of the second-to-last convolutional block of the decoder $\mathcal{D}$. The last convolutional block of $\mathcal{D}$ outputs the final predicted segmentation $\hat{Y}_t$. (Note that networks without decoders can use a structure with a similar purpose, such as a refinement module). The segmentation network loss is a combination of the mask IOU and cross-entropy losses (as in Eq.~\ref{eq:maskflowloss}), encouraging the final segmentation to be close to the ground-truth mask $Y_t$.
See Supplementary for further details.

\subsection{Training}
\label{sec:training}
As the flow module learns only from weakly-supervised signals, it produces distorted flow fields when given unreasonable previous masks. This can occur early in training or during simple joint training of the flow module and segmentation network, when predicted masks are noisy. To overcome these challenges, we use two training mechanisms that stabilize learning: previous mask teacher-forcing
and two-stage training. We analyze hyperparameter robustness and comment further on these two mechanisms in the Supplementary.
\vspace{-1em}
\paragraph{Previous mask teacher-forcing.} Teacher-forcing~\cite{bengio2015scheduled} is a sampling technique widely used in autoregressive model training, where a model's own output is used for its next prediction during inference. As noted in Sec.~\ref{sec:flowmodule}, our flow module warps the previous predicted mask in inference. Thus, we use teacher-forcing in training to allow it to learn good flow fields from warped ground-truth masks as well. In training only, the previous mask $\Tilde{Y}_{t-1}$ warped by the flow module is teacher-forced with the ground-truth mask $Y_{t-1}$ with probability $p$, and the network's previous predicted mask $\hat{Y}_{t-1}$ with probability $1-p$.
\vspace{-1em}
\paragraph{Two-stage training.} We alternately freeze the segmentation model weights to let both modules learn from the progress of the other.~\cite{cheng2017segflow} used a similar strategy when training their joint models. First, we train the segmentation model $\mathcal{S}$ for $E_s$ epochs. We then add the flow module $\mathcal{F}$, which stays unfrozen for the rest of training. We alternately unfreeze and freeze $\mathcal{S}$ every $E_a$ epochs until convergence.

\vspace{-1em}
\section{Experiments}
\vspace{-0.5em}

\subsection{Datasets and Evaluation Metrics}

We train and evaluate on two major benchmark datasets for the VOS task: DAVIS17~\cite{pont-tuset2017davis} and YouTubeVOS~\cite{xu2018youtube}. 
For reproducibility, our code and models will be made available online.

\textbf{DAVIS 17.} DAVIS17~\cite{pont-tuset2017davis} has 120 videos total, with a maximum of 10 objects per video. Using the official dataset splits, we train on the 60 training videos and evaluate on the 30 validation and 30 test-dev videos. We use the official DAVIS17 evaluation protocol~\cite{pont-tuset2017davis}, which is the Jaccard (IOU) mean $\mathcal{J}$ and contour F-score $\mathcal{F}$ across all objects and videos, as well as the mean of $\mathcal{J}$ and $\mathcal{F}$ for the overall score.

\textbf{YouTubeVOS.} YouTubeVOS~\cite{xu2018youtube} is the largest VOS dataset to-date, with a maximum of 12 objects per video. We use the official dataset splits, with 3,471 training videos (training) and 474 validation videos (evaluation). The evaluation protocol also averages $\mathcal{J}$ and $\mathcal{F}$ scores over seen and unseen classes separately, which are then averaged for the overall score.

\vspace{-0.5em}
\subsection{Implementation Details}
\label{sec:implementation}
\textbf{Model.} Our flow module's lightweight U-Net~\cite{ronneberger2015unet} structure uses a ResNet50~\cite{he2016resnet50} encoder pretrained on ImageNet~\cite{deng2009imagenet} and skip-connections between symmetric encoder-decoder blocks. Decoder layers use bilinear upsampling, followed by two $3 \times 3$ convolutions. We integrate our module into the recent state-of-the-art STM-cycle~\cite{li2020stmcycle}.\footnote{\label{fn:baseline}We ran STM-cycle~\cite{li2020stmcycle} using the authors' provided code and could not replicate the reported frame rates in~\cite{li2020stmcycle}. With our TITAN RTX GPU, which is faster than the TITAN Xp in~\cite{li2020stmcycle}, we still only achieved the FPS in the tables. For fair comparison, since~\cite{li2020stmcycle} is the closest work to ours, we report our replicated FPS on the same GPU we used to benchmark our model.~\cite{li2020stmcycle}'s reported FPS corresponding to Tables~\ref{tab:davis17quant}(a),~\ref{tab:davis17quant}(b), and~\ref{tab:davis17quant}(c) are 38, 31, and 43, respectively. Note that the reported YouTubeVOS validation set result in Table 2 of~\cite{li2020stmcycle} also incorrectly switched the STM~\cite{jiang2019stm} $\mathcal{F_S}, \mathcal{J_U}$ scores. We show the correct version from~\cite{jiang2019stm} here in Table~\ref{tab:davis17quant}(c).} Following training procedures in~\cite{li2020stmcycle, jiang2019stm, voigtlaender2019feelvos}, we pool DAVIS17 and YouTubeVOS training splits in all experiments. See Supplementary for training details (e.g. hyperparameters, hardware, and data augmentation).

\subsection{Quantitative Results}

We compare our method's performance and speed against state-of-the-art works on DAVIS17 \cite{pont-tuset2017davis} and YouTubeVOS~\cite{xu2018youtube}. These include works that use extra annotated data (ED), online learning (OL), and optical flow (shown as superscript $F$). Tables~\ref{tab:davis17quant}(a),~\ref{tab:davis17quant}(b), and~\ref{tab:davis17quant}(c) respectively show DAVIS17 validation, DAVIS17 test-dev, and YouTubeVOS validation results.

\begin{table}
\begin{center}
\begin{adjustbox}{width=1.025\linewidth, center}
\setlength\tabcolsep{2.5pt}
\begin{tabular}{lcccccc}
\toprule
\textit{OL \& ED methods} & ED & OL & $\mathcal{J\%}$ & $\mathcal{F\%}$ & $\mathcal{J\&F\%}$ & FPS \\
\midrule
STCNN~\cite{xu2019stcnn} & \checkmark & \checkmark & 58.7 & 64.6 & 61.7 & 0.26$^\dagger$ \\
OnAVOS~\cite{voigtlaender17onavos} & \checkmark & \checkmark & 64.5 & 71.2 & 67.9 & 0.1\\ 
BoLTVOS~\cite{voigtlaender2019boltvos} & \checkmark & \checkmark & 72.0 & 80.6 & 76.3 & 0.69\\
TANDTM~\cite{huang2020template} & \checkmark & \checkmark & 72.3 & 79.4 & 75.9 & 7.1\\
PReMVOS$^F$~\cite{luiten2018premvos} & \checkmark & \checkmark & 73.9 & 81.7 & 77.8 & 0.03 \\ \midrule
OSMN~\cite{yang2018osmn} & \checkmark & - & 52.5 & 57.1 & 54.8 & 8\\ 
RGMP~\cite{oh2018rgmp} & \checkmark & - & 64.8 & 68.6 & 66.7 & 3.6\\ 
AGSS-VOS$^F$~\cite{lin2019agss} & \checkmark & - & 64.9 & 69.9 & 67.4 & 10 \\ 
DMM-Net~\cite{zeng2019dmmnet} & \checkmark & - & 68.1 & 73.3 & 70.7 & -\\
FEELVOS~\cite{voigtlaender2019feelvos} & \checkmark & - & 69.1 & 74.0 & 71.5 & 2\\ 
STM~\cite{jiang2019stm} & \checkmark & - & 79.2 & 84.3 & 81.8 & 6.3\\ \midrule
OSVOS~\cite{caelles2017osvos} & - & \checkmark & 64.7 & 71.3 & 68.0 & 0.1\\
STM-cycle~\cite{li2020stmcycle} & - & \checkmark & 69.3 & 75.3 & 72.3 & 9.3\\[0.25em]
\toprule
\textit{non-OL, non-ED} & ED & OL & $\mathcal{J\%}$ & $\mathcal{F\%}$ & $\mathcal{J\&F\%}$ & FPS\\ \midrule
FAVOS~\cite{cheng2018favos} & - & - & 54.6 & 61.8 & 58.2 & 0.8\\
D3S~\cite{lukezic2020d3s} & - & - & 57.8 & 63.8 & 60.8 & 25\\
STM-cycle~\cite{li2020stmcycle} & - & - & 68.7 & 74.7 & 71.7 & \textbf{31.9}\footnoteref{fn:baseline}\\
\midrule
\textbf{Ours} & - & - & \textbf{70.6} & \textbf{75.8} & \textbf{73.2} & 17.3\\  
\bottomrule
\multicolumn{7}{c}{\rule{0pt}{1.2em}(a)}
\end{tabular}

\begin{tabular}{lcccccc}
\toprule
\textit{OL \& ED methods} & ED & OL & $\mathcal{J\%}$ & $\mathcal{F\%}$ & $\mathcal{J\&F\%}$ & FPS \\
\midrule
OnAVOS~\cite{voigtlaender17onavos} & \checkmark & \checkmark & 53.4 & 59.6 & 56.9 & 0.03\\
TANDTM~\cite{huang2020template} & \checkmark & \checkmark & 61.3 & 70.3 & 65.4 & 7.1\\
PReMVOS$^F$~\cite{luiten2018premvos} & \checkmark & \checkmark & 67.5 & 75.7 & 71.6 & 0.02\\ \midrule
RGMP~\cite{oh2018rgmp} & \checkmark & - & 51.3 & 54.4 & 52.8 & 2.4\\
AGSS-VOS$^F$~\cite{lin2019agss} & \checkmark & - & 54.8 & 59.7 & 57.2 & 9\\
FEELVOS~\cite{voigtlaender2019feelvos} & \checkmark & - & 55.2 & 60.5 & 57.8 & 1.8\\ \midrule
STM-cycle~\cite{li2020stmcycle} & - & \checkmark & 55.3 & 62.0 & 58.6 & 6.9\\[0.25em]
\toprule
\textit{non-OL, non-ED} & ED & OL & $\mathcal{J\%}$ & $\mathcal{F\%}$ & $\mathcal{J\&F\%}$ & FPS\\ \midrule
RVOS~\cite{ventura2019rvos} & - & - & 48.0 & 52.6 & 50.3 & 22.7\\
STM-cycle~\cite{li2020stmcycle} & - & - & 55.1 & 60.5 & 57.8 & \textbf{25.9}\footnoteref{fn:baseline}\\
\midrule
\textbf{Ours} & - & - & \textbf{57.1} & \textbf{63.1} & \textbf{60.1} & 13.7\\  
\bottomrule 
\multicolumn{7}{c}{\rule{0pt}{1.2em}(b)}
\end{tabular}

\begin{tabular}{c|c}\\ & \\ \\ \\ \\ \\ \\ \\ \\ \\ \\ \\ \\ \\ \\ \\ \\ \\ \\ \\ \\ \\ \\ \end{tabular}

\begin{tabular}{lcccccccc}
\toprule
\textit{OL \& ED methods} & ED & OL & $\mathcal{G\%}$ &  $\mathcal{J_S\%}$ & $\mathcal{J_U\%}$ & $\mathcal{F_S\%}$ & $\mathcal{F_U\%}$ & FPS \\
\midrule
MaskTrack~\cite{perazzi2017learning} & \checkmark & \checkmark & 53.1 & 59.9 & 45.0 & 59.5 & 47.9 & 0.05\\
OnAVOS~\cite{voigtlaender17onavos} & \checkmark & \checkmark & 55.2 & 60.1 & 46.6 & 62.7 & 51.4 & 0.05\\
DMM-Net~\cite{zeng2019dmmnet} & \checkmark & \checkmark & 58.0 & 60.3 & 50.6 & 63.5 & 57.4 & -\\
PReMVOS$^F$~\cite{luiten2018premvos} & \checkmark & \checkmark & 66.9 & 71.4 & 56.5 & 75.9 & 63.7 & 0.17\\
BoLTVOS~\cite{voigtlaender2019boltvos} & \checkmark & \checkmark & 71.1 & 71.6 & 64.3 & - & - & 0.74\\ \midrule
OSMN~\cite{yang2018osmn} & \checkmark & - & 51.2 & 60.0 & 40.6 & 60.1 & 44.0 & 4.2\\
DMM-Net~\cite{zeng2019dmmnet} & \checkmark & - & 51.7 & 58.3 & 41.6 & 60.7 & 46.3 & 12\\
RGMP~\cite{oh2018rgmp} & \checkmark & - & 53.8 & 59.5 & - & 45.2 & - & 7\\
AGSS-VOS$^F$~\cite{lin2019agss} & \checkmark & - & 71.3 & 71.3 & 65.5 & 75.2 & 73.1 & 12.5\\
STM~\cite{jiang2019stm} & \checkmark & - & 79.4 & 79.7 & 72.8 & 84.2 & 80.9 & 6.3 \\ \midrule
OSVOS~\cite{caelles2017osvos} & - & \checkmark & 58.8 & 59.8 & 54.2 & 60.5 & 60.7 & 0.06\\
S2S~\cite{xu2018youtube} & - & \checkmark & 64.4 & 71.0 & 55.5 & 70.0 & 61.2 & 0.06\\
STM-cycle~\cite{li2020stmcycle} & - & \checkmark & 70.8 & 72.2 & 62.8 & 76.3 & 71.9 & 13.8\\[0.25em]
\toprule
\textit{non-OL, non-ED} & ED & OL & $\mathcal{G\%}$ & $\mathcal{J_S\%}$ & $\mathcal{J_U\%}$ & $\mathcal{F_S\%}$ & $\mathcal{F_U\%}$ & FPS \\ \midrule
RVOS~\cite{ventura2019rvos} & - & - & 56.8 & 63.6 & 45.5 & 67.2 & 51.0 & 24\\
S2S~\cite{xu2018youtube} & - & - & 57.6 & 66.7 & 48.2 & 65.5 & 50.3 & 6\\
STM-cycle~\cite{li2020stmcycle} & - & - & 69.9 & \textbf{71.7} & 61.4 & \textbf{75.8} & 70.4 & \textbf{30.3}\footnoteref{fn:baseline}\\
\midrule
\textbf{Ours} & - & - & \textbf{71.1} & \textbf{71.7} & \textbf{64.0} & 75.2 & \textbf{73.3} & 16.7\\  
\bottomrule
\multicolumn{9}{c}{\rule{0pt}{1.2em}(c)}
\end{tabular}
\end{adjustbox}
\end{center}
\vspace{-0.5em}
\caption{Comparison with state-of-the-art methods on DAVIS17 validation (a), DAVIS17 test-dev (b), and YouTubeVOS validation (c). `ED' denotes usage of extra training data. `OL' denotes online learning. Superscript `$F$' denotes usage of optical flow. In (a), $\dagger$ denotes runtimes only available on DAVIS16. In (c), $\mathcal{S}$, $\mathcal{U}$ subscripts denote classes seen and unseen in training, and $\mathcal{G}$ is the global mean. Other method results in (c) taken from~\cite{li2020stmcycle, jiang2019stm}.
\vspace{-1em}
}
\label{tab:davis17quant}
\end{table}

\textbf{DAVIS17.} As shown in Tables~\ref{tab:davis17quant}(a) and~\ref{tab:davis17quant}(b), on the official validation and test-dev splits, we achieve state-of-the-art performance among methods that do not use extra data (ED) nor online learning (OL) (and many that do), 
while maintaining high frame rates. We achieve +1.9\% and +2.0\% $\mathcal{J}$ gains over~\cite{li2020stmcycle} respectively. 
Notably, on both splits, we outperform STM-cycle~\cite{li2020stmcycle}'s online learning version, even though it optimizes for performance by adding time-costly iterative mask refinement. This demonstrates that our flow module's warped masks can replace detailed mask refinement without sacrificing speed. Among works that use optical flow, on validation, we achieve a +5.8\% $\mathcal{J}\&\mathcal{F}$ improvement over AGSS-VOS~\cite{lin2019agss} and stay competitive with PReMVOS~\cite{luiten2018premvos} with significantly faster speeds.

\textbf{YouTubeVOS.} YouTubeVOS is the largest VOS benchmark. Since it evaluates performance on classes unseen in training, it measures generalization well. As shown in Table~\ref{tab:davis17quant}(c), on the official validation set, we achieve state-of-the-art performance among methods that do not use extra data (ED) nor online learning (OL), while maintaining a high frame rate. We also outperform all but two works that use ED, OL, or both, which both have lower frame rates. Crucially, our method generalizes significantly better to unseen classes than both offline (+2.6\% $\mathcal{J_U}$, +2.9\% $\mathcal{F_U}$) and online versions of STM-cycle~\cite{li2020stmcycle}, showing that our foreground-targeted approach learns motion priors to better segment unseen objects. We outperform the optical flow-equipped PReMVOS~\cite{luiten2018premvos} by +7.5\% $\mathcal{J_U}$ with 98 times the speed, highlighting the strengths of our visual warping compared to traditional optical flow.

\begin{figure*}
\begin{center}
\setlength\tabcolsep{0.85pt}
\begin{tabular}{ccc|ccc}
    \includegraphics[width=0.1633\linewidth]{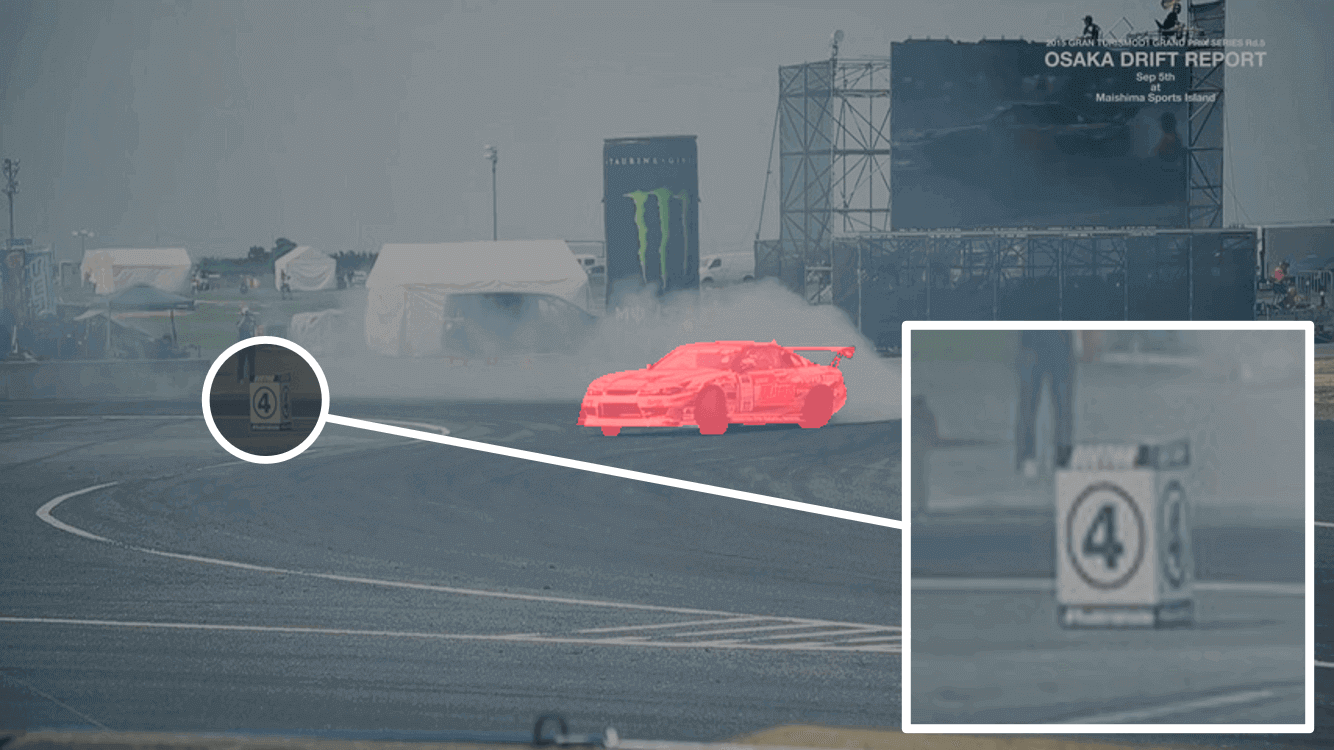} & \includegraphics[width=0.1633\linewidth]{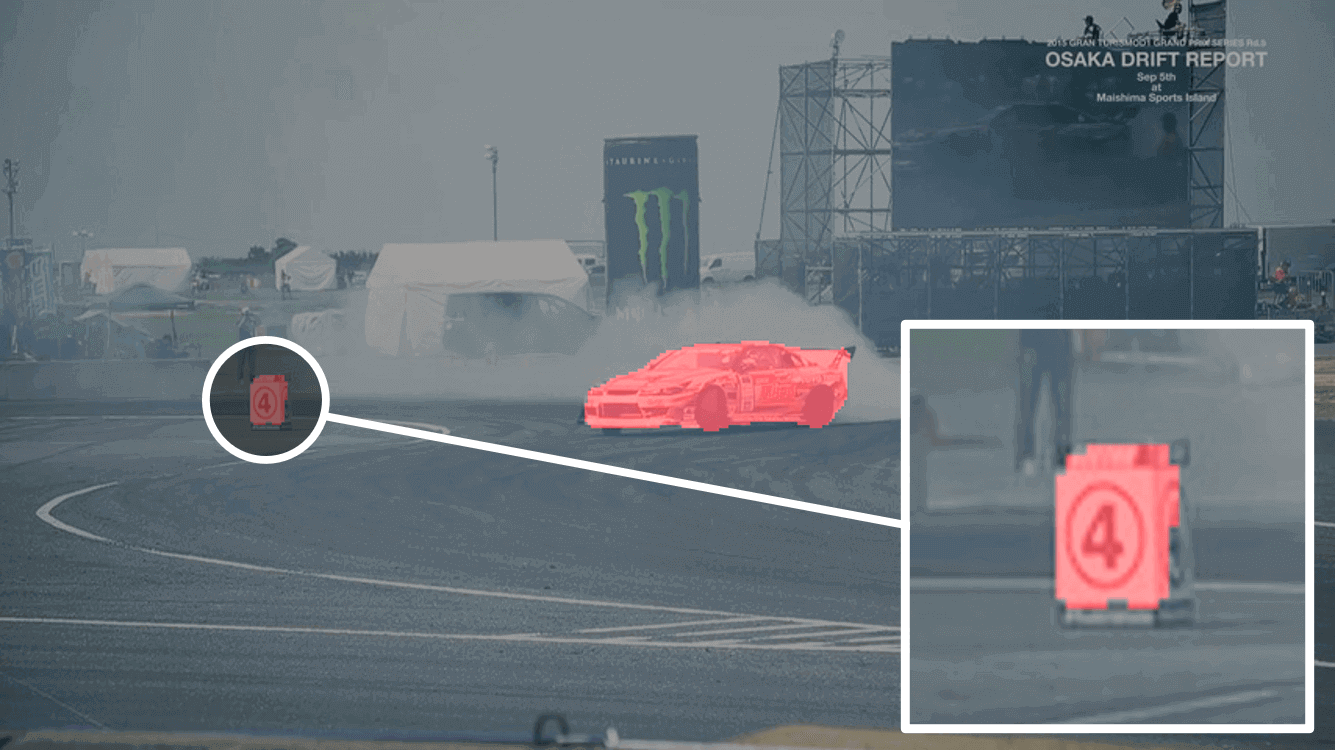} & \includegraphics[width=0.1633\linewidth]{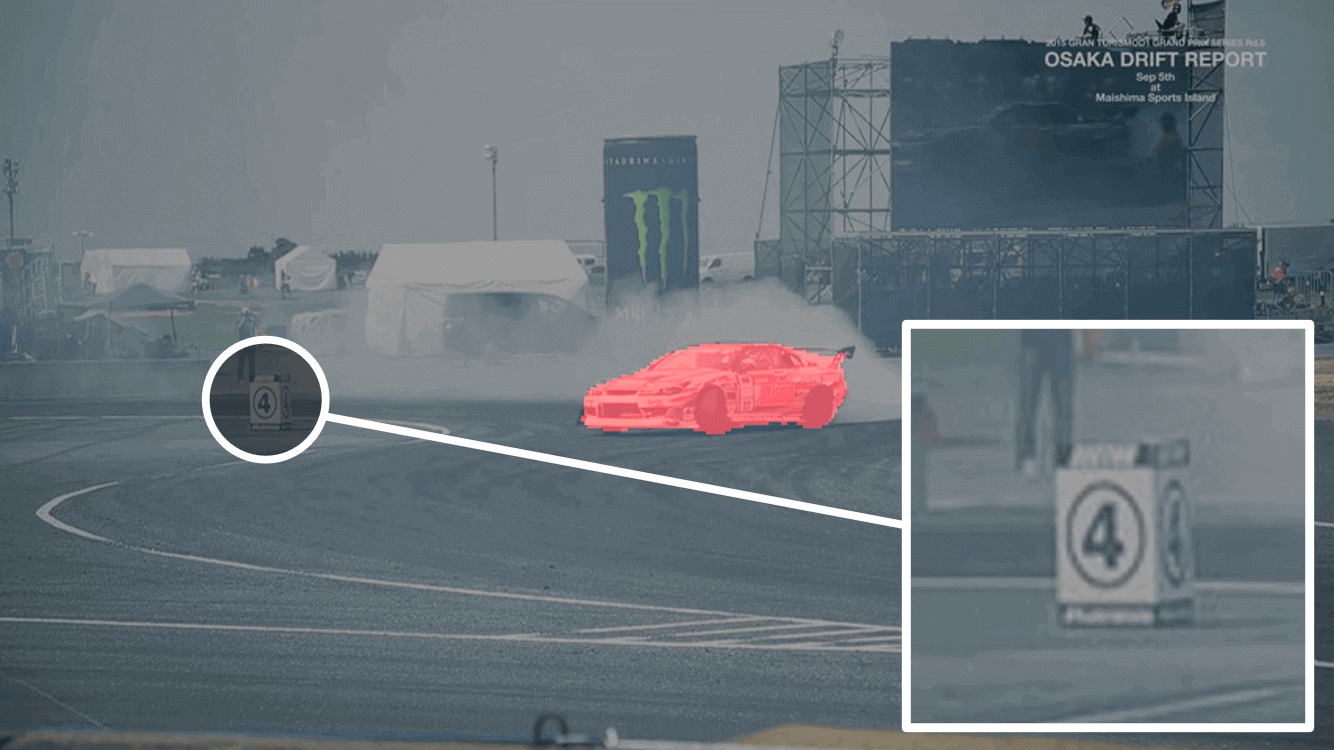} & \includegraphics[width=0.1633\linewidth]{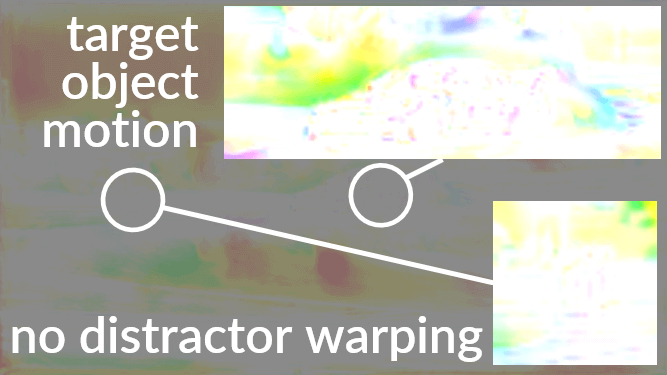} & \includegraphics[width=0.1633\linewidth]{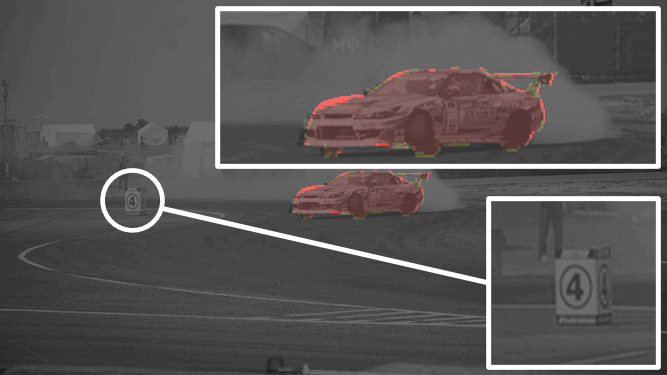} & \includegraphics[width=0.1633\linewidth]{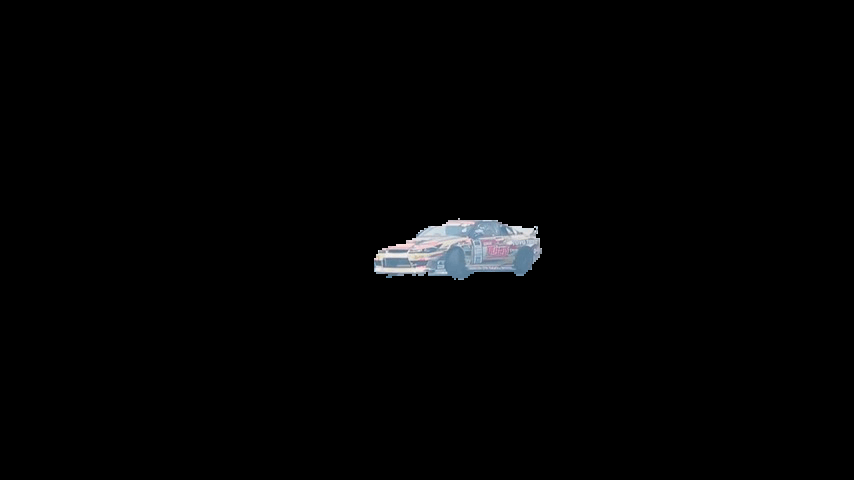} \\
    \includegraphics[width=0.1633\linewidth]{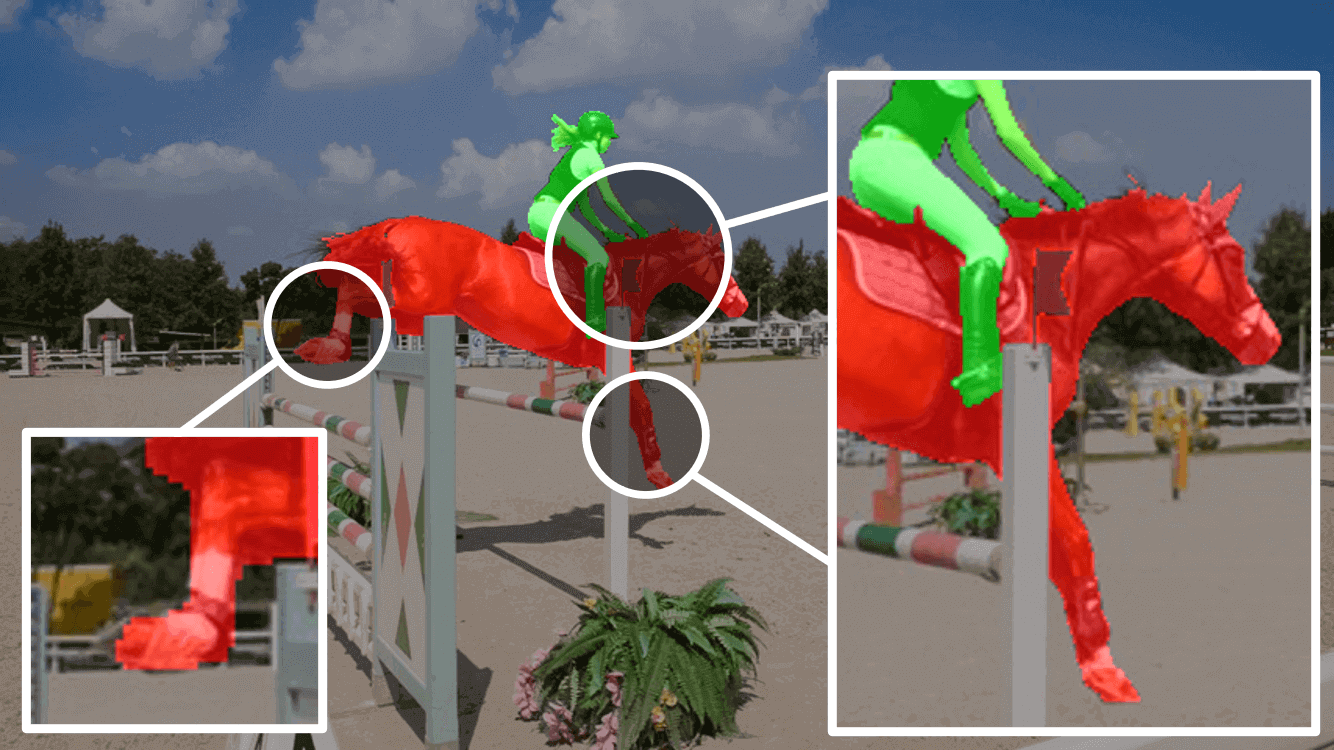} & \includegraphics[width=0.1633\linewidth]{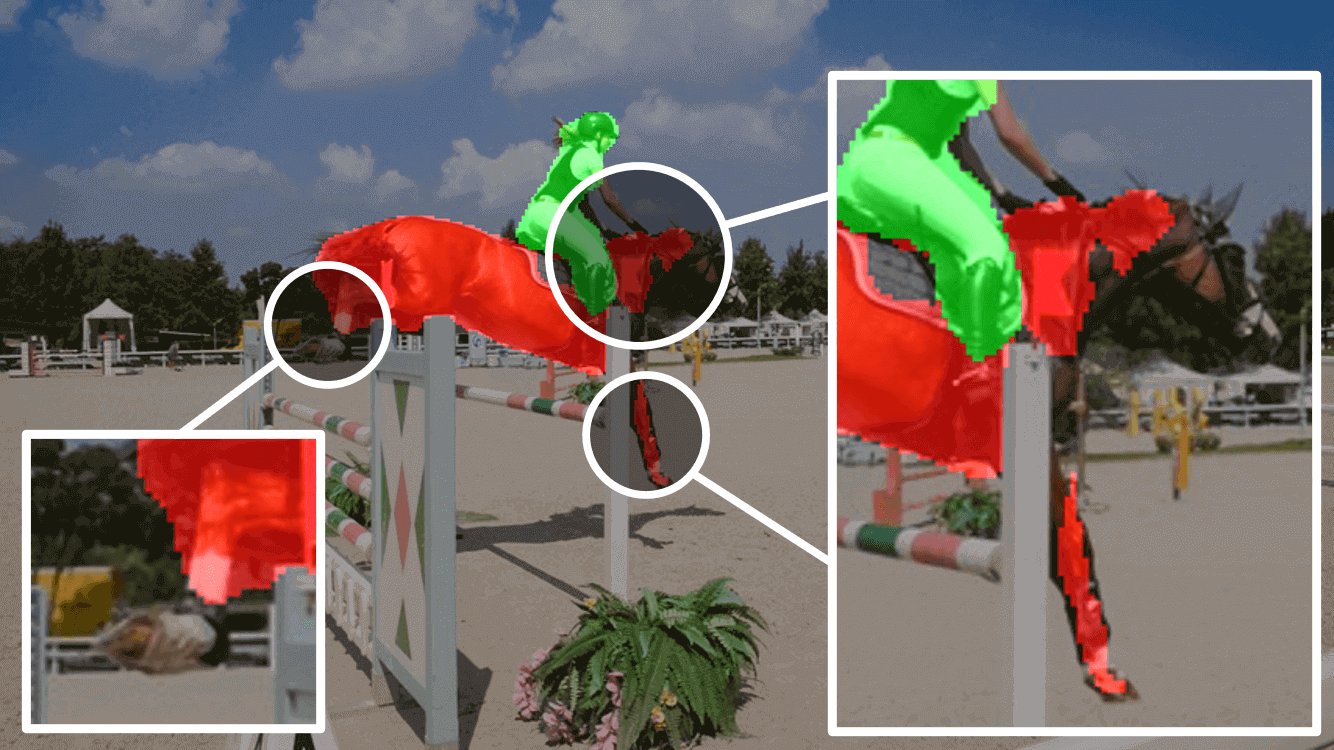} & \includegraphics[width=0.1633\linewidth]{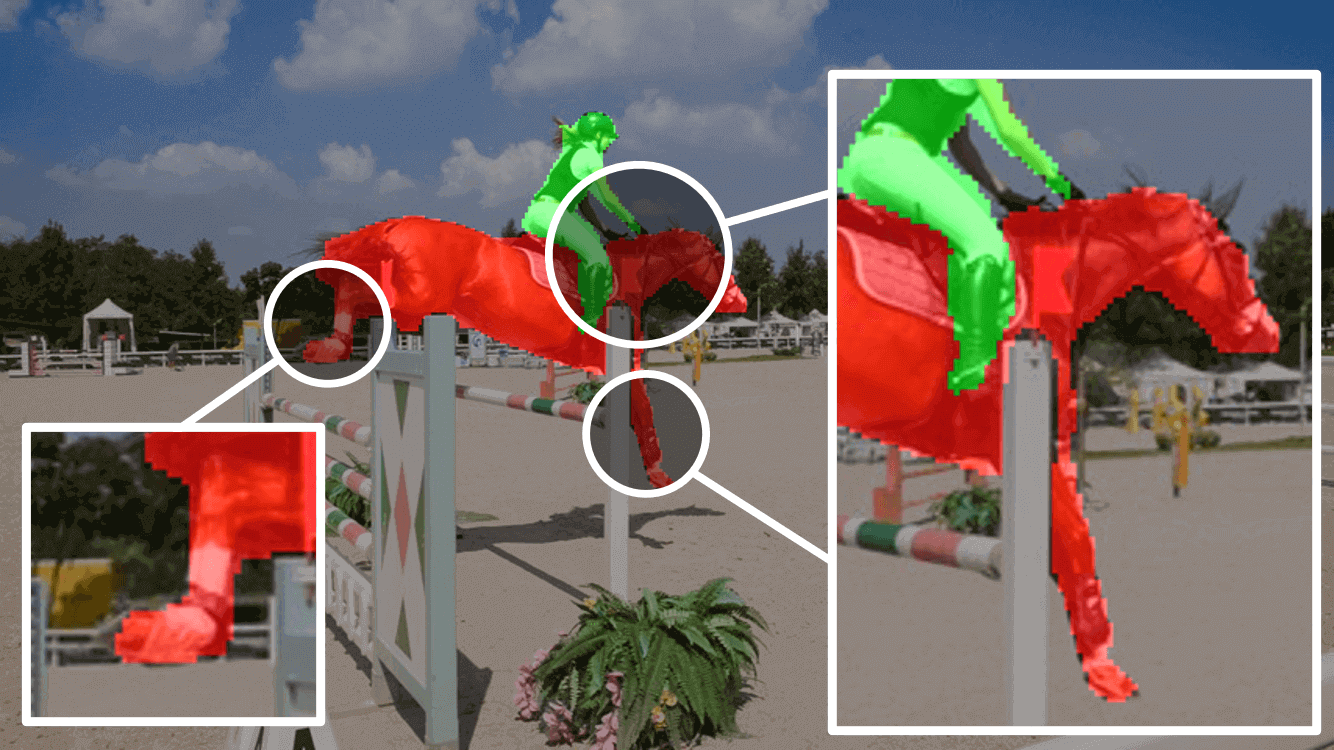} & \includegraphics[width=0.1633\linewidth]{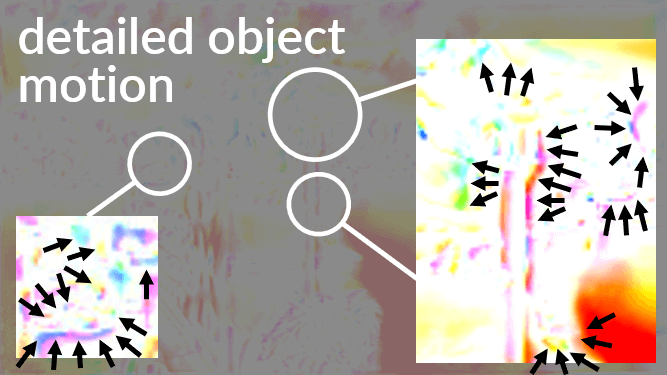} & \includegraphics[width=0.1633\linewidth]{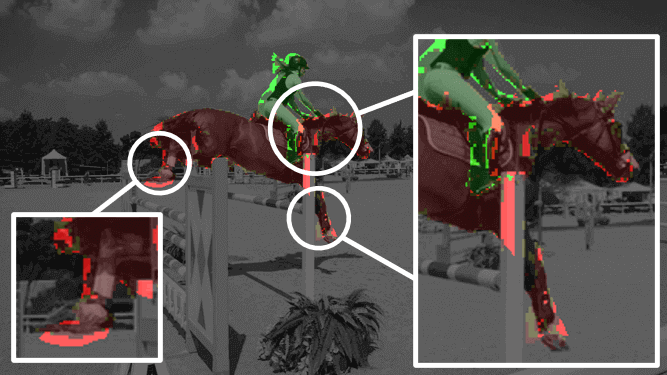} & \includegraphics[width=0.1633\linewidth]{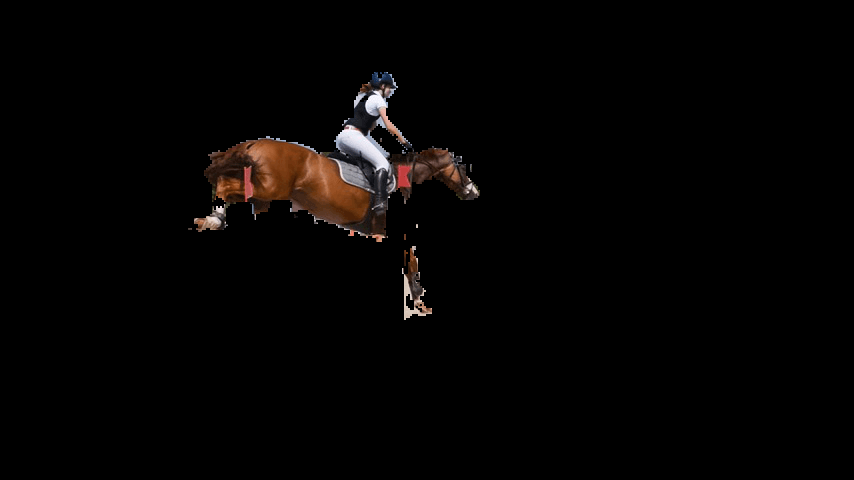} \\
    \includegraphics[width=0.1633\linewidth]{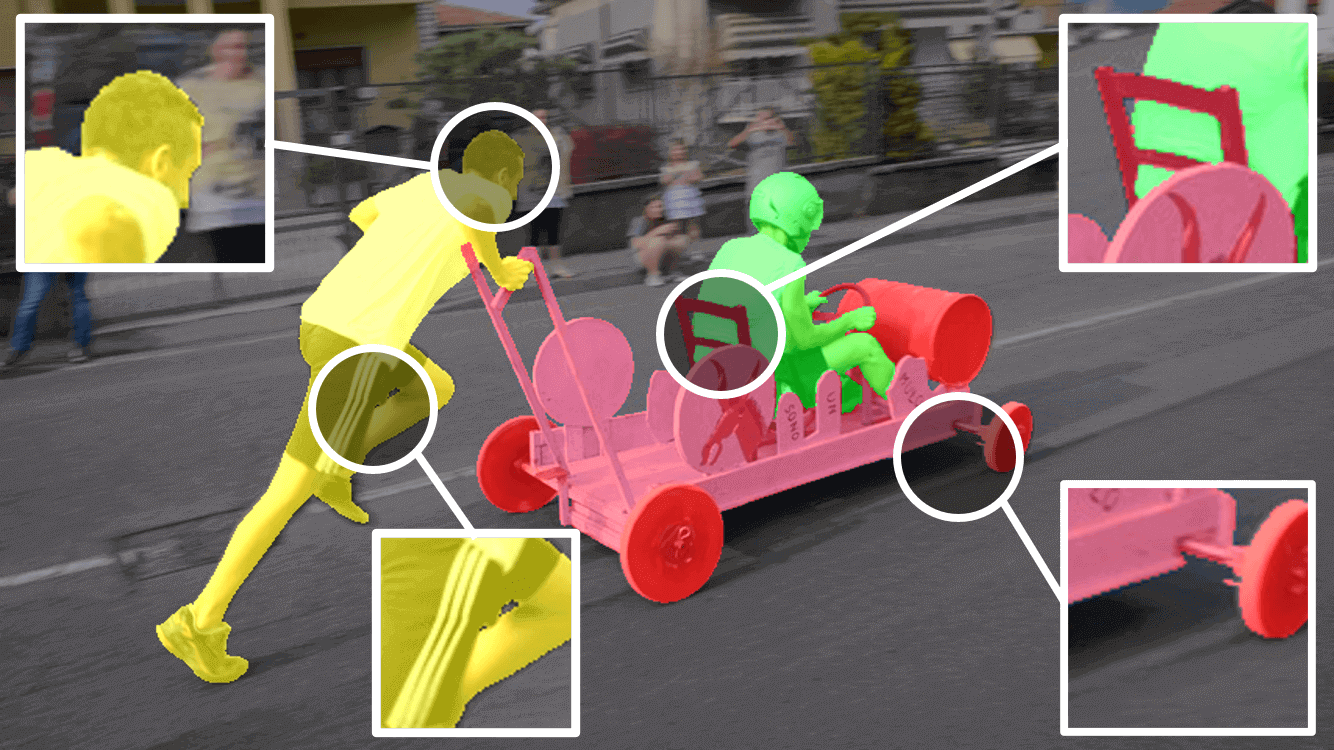} & \includegraphics[width=0.1633\linewidth]{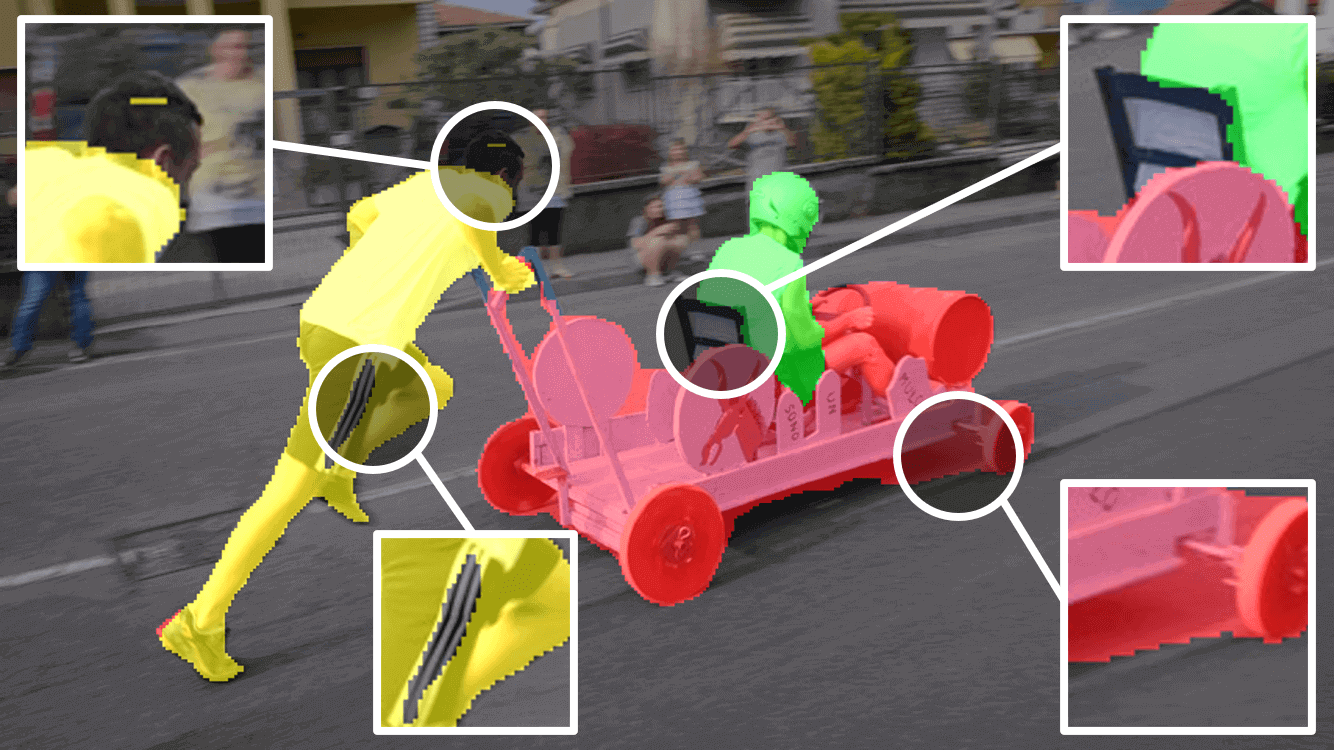} & \includegraphics[width=0.1633\linewidth]{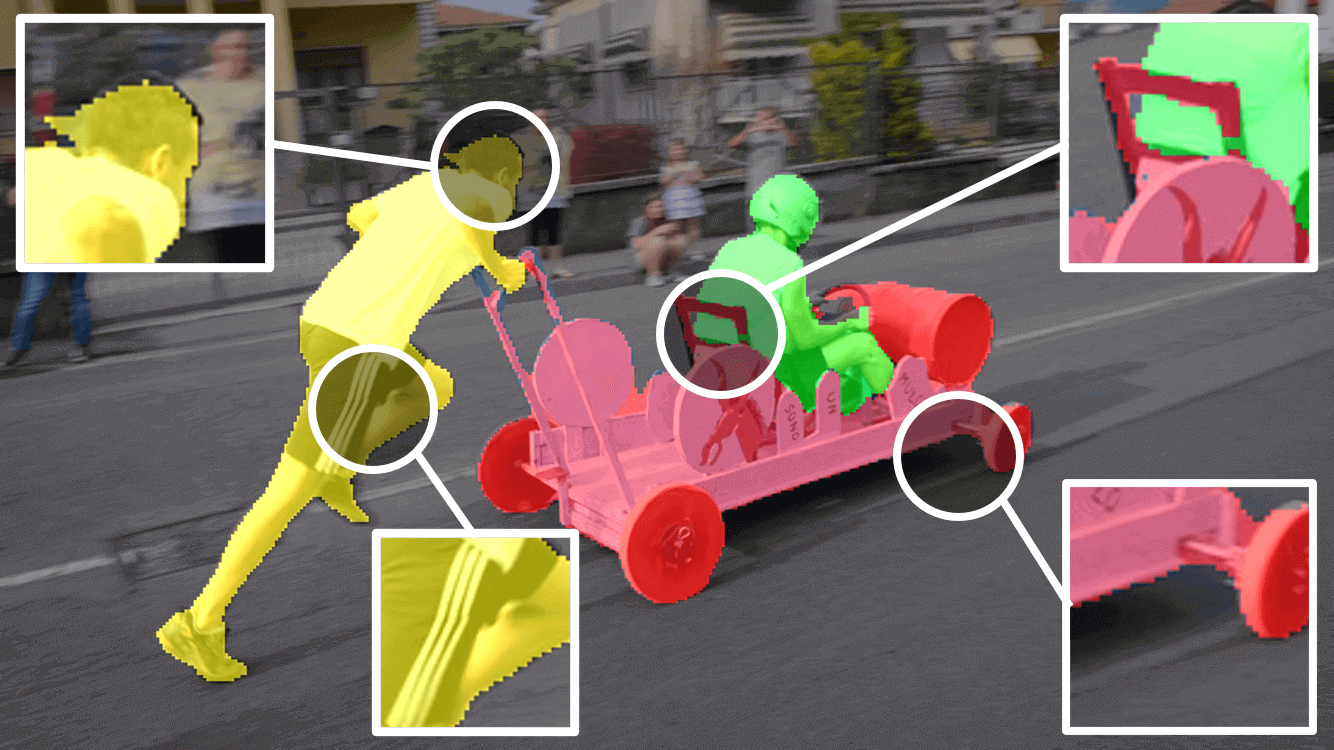} & \includegraphics[width=0.1633\linewidth]{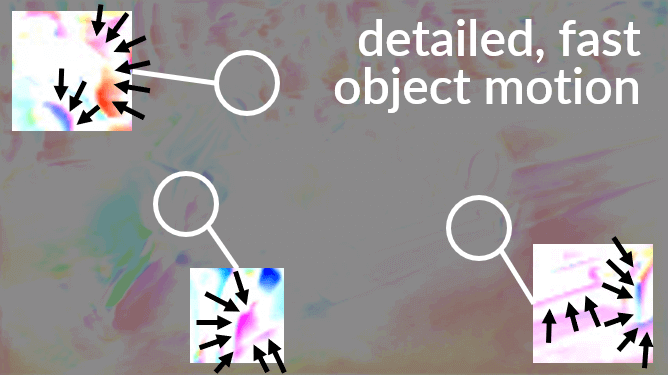} & \includegraphics[width=0.1633\linewidth]{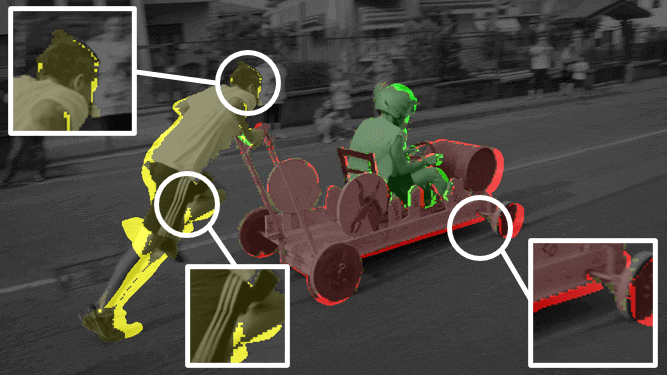} & \includegraphics[width=0.1633\linewidth]{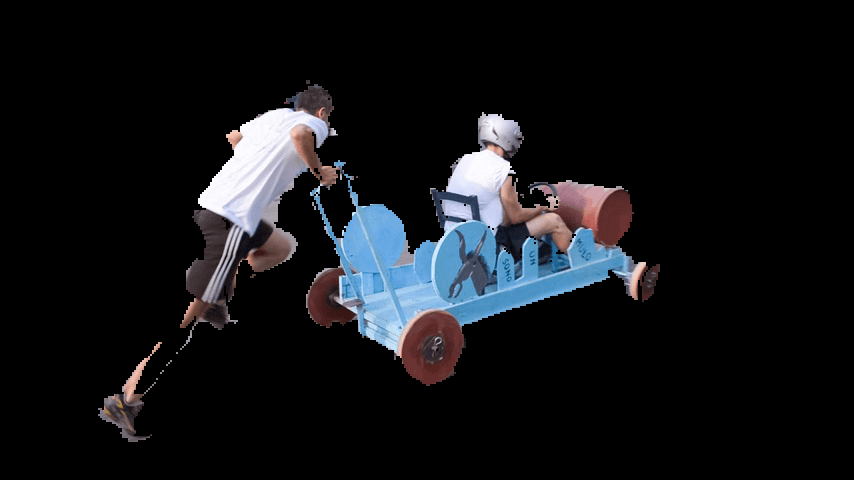} \\
    \scriptsize (a) Ground-truth & \scriptsize (b) STM-cycle~\cite{li2020stmcycle} & \scriptsize (c) Ours & \scriptsize (c1) Flow field & \scriptsize (c2) Warp diff. & \scriptsize (c3) Warped frame \\
    \vspace{-2em}
\end{tabular}
\end{center}
  \caption{Qualitative comparison with STM-cycle~\cite{li2020stmcycle} on DAVIS17 validation (a, b, c), and our method's intermediate outputs (c1-3). Following~\cite{ilg2017flownet2}, we color-code flow fields (c1) with polar coordinate displacements. (c2) brightens pixels that exist in the previous, but not the warped mask, highlighting motion that corresponds to the flows. The masked warped frames (c3) show that our warping operation accurately preserves object detail.
  In row 1, our flow (c1) captures the target car's leftward (green, blue) motion, but not the distractor due to the foreground-focused losses. In row 2, our detailed flow field propagates object details, such as warping the back hoof upward (purple). 
  In row 3, even with fast-moving objects, the flow accurately warps boundaries of details like the lower right wheel and upper left head.
  }
\label{fig:qualcompare}
\vspace{-1em}
\end{figure*}

\subsection{Ablation Analysis}
In the top half of Table~\ref{tab:ablationquant}, we analyze relative contributions of key components of our method. 
Our score drops 1.0\% with either just the Mask Flow Loss or Visual Flow Loss (VFL); this shows the benefit of leveraging both warped masks and frames in our method. Without masking the warped frame in the VFL, our score is similar to not using the VFL at all, showing the importance of masking the foreground to eliminate background noise, such as the distractors in the first row of Figure~\ref{fig:qualcompare}. In the bottom half of Table~\ref{tab:ablationquant}, we show the importance of using the warped previous mask for mask refinement. When we replace the input to the decoder with the previous predicted mask or video frame, the network performs on-par with STM-cycle~\cite{li2020stmcycle}, meaning these inputs provide less useful information for refining the mask. With the flow field, performance still lacks by 0.6\%, showing the benefit of explicitly warping the previous mask. The previous masked warped frame expectedly performs on-par with our method, since we warp it identically to the previous mask; still, this shows that the warped mask provides stronger signal about object motion.
\begin{table}
\begin{center}
\begin{adjustbox}{width=0.47\linewidth}
\setlength\tabcolsep{5pt}
\begin{tabular}{llc}
\toprule
\textit{\textbf{Ablations}} & & $\mathcal{J\&F\%}$ \\
\midrule
\multicolumn{2}{l}{STM~\cite{jiang2019stm} (same train protocol as~\cite{li2020stmcycle}, no extra data)} & 70.5 \\
STM-cycle~\cite{li2020stmcycle} & $+$Cycle consistency loss & 71.7 \\
\midrule
Weakly-supervised & $+$Mask Flow Loss only & 72.2 \\
flow module losses & $+$Visual Flow Loss only & 72.2 \\
\midrule
Foreground masking & VFL w/o foreground masking & 72.4 \\
\bottomrule
\multicolumn{3}{l}{\rule{0pt}{1.5em}\textit{\textbf{Alternative inputs to segmentation decoder}}}\\
\toprule
\multicolumn{2}{l}{Previous predicted mask} & 71.6 \\
\multicolumn{2}{l}{Previous video frame} & 71.9 \\
\multicolumn{2}{l}{Flow field} & 72.6 \\
\multicolumn{2}{l}{Previous masked warped frame} & 73.0 \\
\midrule \midrule  
\textbf{Ours} & & \textbf{73.2} \\
\bottomrule 
\end{tabular}
\end{adjustbox}
\end{center}
\caption{Ablation study of our method components on DAVIS17 validation. On top, we show ablations using only the Mask Flow Loss (MFL) or Visual Flow Loss (VFL), and without foreground masking of the VFL. The bottom shows alternative segmentation decoder inputs instead of the warped previous mask that our method uses.
}
\label{tab:ablationquant}
\vspace{-0.5em}
\end{table}

\subsection{Qualitative Results}
\vspace{-0.3em}
In contrast to prior works, we show that our foreground-targeted approach for VOS-specific visual warping produces detailed flow fields without needing to learn the general optical flow task. Figure~\ref{fig:qualcompare} illustrates our method's improvements over the state-of-the-art STM-cycle~\cite{li2020stmcycle} in segmentation detail (see Supplementary figures for temporal consistency).

In Figure~\ref{fig:qualcompare}, we show that our method preserves segmentation detail in cases with background distractors (row 1), small object details (2), and fast-moving objects (3). Notice that our flow fields (c1) are detailed and correspond to the pixel-wise differences between the previous and warped mask highlighted in (c2). Note that since we do not learn general optical flow (which has ground-truth), there is more than one possible flow field that can accurately warp a foreground object in our VOS-specific setting. This means we can learn detailed motion with greater flexibility;
for instance, our model often warps background pixels to achieve better foreground alignment (e.g. the car boundary in row 1), which traditional optical flow would penalize, but our weakly-supervised losses do not. 
Our visual warping method also enables stronger temporal consistency despite diverse challenges (see Supplementary).

\vspace{-1em}
\section{Conclusion}
\vspace{-0.5em}

We propose a novel foreground-targeted visual warping approach that improves segmentation detail and temporal consistency for semi-supervised video object segmentation (VOS). Instead of learning full optical flow, our flow module learns detailed flow fields using two weakly-supervised losses that directly leverage the target VOS data, which could benefit diverse use cases. We show that the resulting warped masks from our method effectively refine the final segmentations.
Our module can be easily integrated into state-of-the-art segmentation networks. Since it does not predict full optical flow, it is lightweight, fast, and requires no extra training data. On the DAVIS17 and YouTubeVOS benchmarks, we achieve state-of-the-art performance among methods that do not use online learning nor extra data. We also outperform or stay competitive with those that do, while maintaining faster frame rates.

\section*{Acknowledgements}
This work was partially supported by a grant from the Isackson Family Fund for Research in Head and Neck Surgery for compute and support for authors J.G. and F.C.H. The authors also thank Joy Hsu, Shih-Cheng Huang, Rui Yan, Jeffrey Gu, Ali Mottaghi, Nikita Bedi, Danfei Xu, Geeticka Chauhan, and Benjamin Newman for their thoughts, suggestions, and support.

\bibliography{bib}

\end{document}